\documentclass[11pt,a4paper]{article}
\usepackage[hyperref]{emnlp-ijcnlp-2019}
\usepackage{times}
\usepackage{latexsym}

\usepackage{url}

\usepackage{graphicx}   
\usepackage{subfigure}
\usepackage{multirow}
\usepackage{diagbox}
\usepackage{color}
\usepackage{bbm}
\usepackage{multicol}
\usepackage{blindtext}
\usepackage{amsmath,amsfonts,amssymb}
\usepackage{bm}
\usepackage{mathtools}
\usepackage{wrapfig}
\usepackage{wrapfig,lipsum,booktabs}
\usepackage{algorithm}
\usepackage{algpseudocode}

\DeclareMathOperator*{\softmax}{softmax}

\DeclareMathOperator*{\multihead}{MultiHead}
\DeclareMathOperator*{\ffn}{FFN}

\DeclareMathOperator*{\gelu}{Gelu}

\DeclareMathOperator*{\Afind}{find}
\DeclareMathOperator*{\Acount}{count}
\DeclareMathOperator*{\Ain}{in}
\DeclareMathOperator*{\Aunion}{union}
\DeclareMathOperator*{\Ainter}{inter}
\DeclareMathOperator*{\Adiff}{diff}
\DeclareMathOperator*{\Alarge}{large}
\DeclareMathOperator*{\Aless}{less}
\DeclareMathOperator*{\Aequal}{equal}
\DeclareMathOperator*{\Aargmax}{argmax}
\DeclareMathOperator*{\Aargmin}{argmin}
\DeclareMathOperator*{\Afilter}{filter}
\DeclareMathOperator*{\Aset}{set}

\graphicspath{{./figs/}}

\aclfinalcopy 

\title{Multi-Task Learning for Conversational Question Answering \\ over a Large-Scale Knowledge Base}

\author{
	Tao Shen$^{1}$\thanks{~~~Work done while the author was an intern at Microsoft, Beijing, China. }, 
	Xiubo Geng$^2$, Tao Qin$^2$, Daya Guo$^3$, Duyu Tang$^2$, Nan Duan$^2$, \\
	\textbf{Guodong Long$^1$ and Daxin Jiang$^2$} \\
	$^1$Centre for AI, School of Computer Science, FEIT, University of Technology Sydney \\
	$^2$Microsoft, Beijing, China \\
	$^3$The School of Data and Computer Science, Sun Yat-sen University\\
    {\tt tao.shen@student.uts.edu.au},~~{\tt guodong.long@uts.edu.au} \\
    {\tt \{xiubo.geng,taoqin,dutang,nanduan,djiang\}@microsoft.com} \\
    {\tt guody5@mail2.sysu.edu.cn} 
}

\date{}

\begin{document}
\maketitle
\begin{abstract}
	We consider the problem of conversational question answering over a \textit{large-scale} knowledge base. To handle huge entity vocabulary of a large-scale knowledge base, recent neural semantic parsing based approaches usually decompose the task into several subtasks and then solve them sequentially, which leads to following issues: 1) errors in earlier subtasks will be propagated and negatively affect downstream ones; and 2) each subtask cannot naturally share supervision signals with others. To tackle these issues, we propose an innovative multi-task learning framework where a pointer-equipped semantic parsing model is designed to resolve coreference in conversations, and naturally empower joint learning with a novel type-aware entity detection model. The proposed framework thus enables shared supervisions and alleviates the effect of error propagation. Experiments on a large-scale conversational question answering dataset containing 1.6M question answering pairs over 12.8M entities show that the proposed framework improves overall F1 score from 67\% to 79\% compared with previous state-of-the-art work. 
\end{abstract}

\section{Introduction}

Recent decades have seen the development of AI-driven personal assistants (e.g., Siri, Alexa, Cortana, and Google Now) that often need to answer factorial questions. Meanwhile, large-scale knowledge base (KB) like DBPedia \cite{auer2007dbpedia} or Freebase \cite{bollacker2008freebase} has been built to store world's facts in a structure database, which is used to support open-domain question answering (QA) in those assistants.

Neural semantic parsing based approach \cite{jia2016recombination, reddy2014large,dong2016, liang2016neural, dong2018,d2a} is gaining rising attention for knowledge-based question answer (KB-QA) in recent years since it does not rely on hand-crafted features and is easy to adapt across domains. Traditional approaches usually retrieve answers from a small KB (e.g., small table) \cite{jia2016recombination, xiao2016sequence} and are difficult to handle large-scale KBs. Many recent neural semantic parsing based approaches for KB-QA take a stepwise framework to handle this issue. For example, \citet{liang2016neural}, \citet{dong2016}, and \citet{d2a} first use an entity linking system to find entities in a question, and then learn a model to map the question to logical form based on that. \citet{dong2018} decompose the semantic parsing process into two stages. They first generate a rough sketch of logical form based on low-level features, and then fill in missing details by considering both the question and the sketch. 

However, these stepwise approaches have two issues. First, errors in upstream subtasks (e.g., entity detection and linking, relation classification) are propagated to downstream ones (e.g., semantic parsing), resulting in accumulated errors. For example, case studies in previous works \cite{stagg,dong2016,xu2016question,d2a} show that entity linking error is one of the major errors leading to wrong predictions in KB-QA. Second, since models for the subtasks are learned independently, the supervision signals cannot be shared among the models for mutual benefits. 

\begin{table*}[t]\small
	\centering
	\begin{tabular}{l|l|l}
		\hline
		Alias & Operator & Comments \\ \hline
		A1/2/3&            $start \rightarrow set/num/bool$&          \\
		A4&          $set \rightarrow \Afind(set, p)$&          set of entities with a predicate $p$ edge to entity $e$\\
		A5&          $num \rightarrow \Acount(set)$&          number of distinct elements in the input $set$\\
		A6&          $bool \rightarrow \Ain(e, set)$&         whether the entity $e$ in $set$ or not\\
		A7&          $set \rightarrow \Aunion(set_1, set_2)$ &          $set_1 \cup set_2$\\
		A8&          $set \rightarrow \Ainter(set_1, set_2)$&          $set_1 \cap set_2$\\
		A9&          $set \rightarrow \Adiff(set_1, set_2)$&          $set_1$ - $set_2$\\
		A10&          $set \rightarrow \Alarge(set, p, num)$&          subset of set linking to more than $num$ entities with predicate $p$\\
		A11&          $set \rightarrow \Aless(set, p, num)$&          subset of set linking to less than $num$ entities with predicate $p$\\
		A12&          $set \rightarrow \Aequal(set, p, num)$&          subset of set linking to $num$ entities with predicate $p$\\
		A13&          $set \rightarrow \Aargmax(set, p)$&          subset of set linking to most entities with predicate $p$\\
		A14&          $set \rightarrow \Aargmin(set, p)$&          subset of set linking to least entities with predicate $p$\\
		A15&          $set \rightarrow  \Afilter(tp, set) $&   subset where entity $e$ in set and belong to entity type $tp$   \\	
		A16&          $num \rightarrow u\_num$&    transform number in utterance $u\_num$ to intermediate number $num$  \\ 
		A17&          $set \rightarrow \Aset(e)$&           \\		
		A18/19/20/21 &          $e/p/tp/u\_num \rightarrow$ constant&  *instantiation for $e$, $p$, $tp$, $u\_num$       \\	
		\hline
	\end{tabular}
	\caption{Brief grammar definitions for logical form generation. *instantiation of entity $e$, predicate $p$, type $tp$, number-in-question $u\_num$, by corresponding constant parsed from the question.}
	\label{tb:csqa_grammer_full}
\end{table*}

To tackle issues mentioned above, we propose a novel multi-task semantic parsing framework for KB-QA. Specifically, an innovative pointer-equipped semantic parsing model is first designed for two purposes: 1) built-in pointer network toward positions of entity mentions in the question can naturally empower multi-task learning with conjunction of upstream sequence labeling subtask, i.e., entity detection; and 2) it explicitly takes into account the context of entity mentions by using the supervision of the pointer network. Besides, a type-aware entity detection method is proposed to produce accurate entity linking results, in which, a joint prediction space combining entity detection and entity type is employed, and the predicted type is then used to filter entity linking results during inference phase. 

The proposed framework has certain merits. 

First, since the two subtasks, i.e., pointer-equipped semantic parsing and entity detection, are closely related, learning them within a single model simultaneously makes the best of supervisions and improves performance of KB-QA task.

Second, considering entity type prediction is crucial for entity linking, our joint learning framework combining entity mention detection with type prediction  leverages contextual information, and thus further reduces errors in entity linking.

Third, our approach is naturally beneficial to coreference resolution for conversational QA due to rich contextual features captured for entity mention, compared to previous works directly employing low-level features (e.g., mean-pooling over word embeddings) as the representation of an entity. This is verified via our experiments in \S\ref{sec:exp_model_comp}. 

We evaluate the proposed framework on the CSQA \cite{csqa} dataset, which is the largest public dataset for complex conversational question answering over a large-scale knowledge base. Experimental results show that the overall F1 score is improved by 12.56\% compared with strong baselines, and the improvements are consistent for all question types in the dataset.

\section{Task Definition} \label{sec:sp_prereq}

In this work, we target the problem of conversational question answering over a large-scale knowledge base. Formally, in training data, question $U$ denotes an user utterance from a dialog, which is concatenated dialog history for handling ellipsis or coreference in conversations, and the question is labeled with its answer $A$. Besides, ``IOB'' (Insider-Outside-Beginning) tagging and entities linking to KB are also labeled for entity mentions in $U$  to train an entity detection model. 

We employ a neural semantic parsing based approach to tackle the problem. That is, given a question, a semantic parsing model is used to produce a logical form which is then executed on the KB to retrieve an answer. We decompose the approach into two subtasks, i.e., entity detection for entity linking and semantic parsing for logical form generation. The former employs IOB tagging and corresponding entities as supervision, while the latter uses a gold logical form as supervision, which may be obtained by conducting intensive BFS\footnote{Breadth-first search with limited buffer \cite{d2a}} over KB if only final answers (i.e., weak supervision) are provided. 

\begin{figure*}[htbp] 
	\centering
	\includegraphics[width=0.999\textwidth]{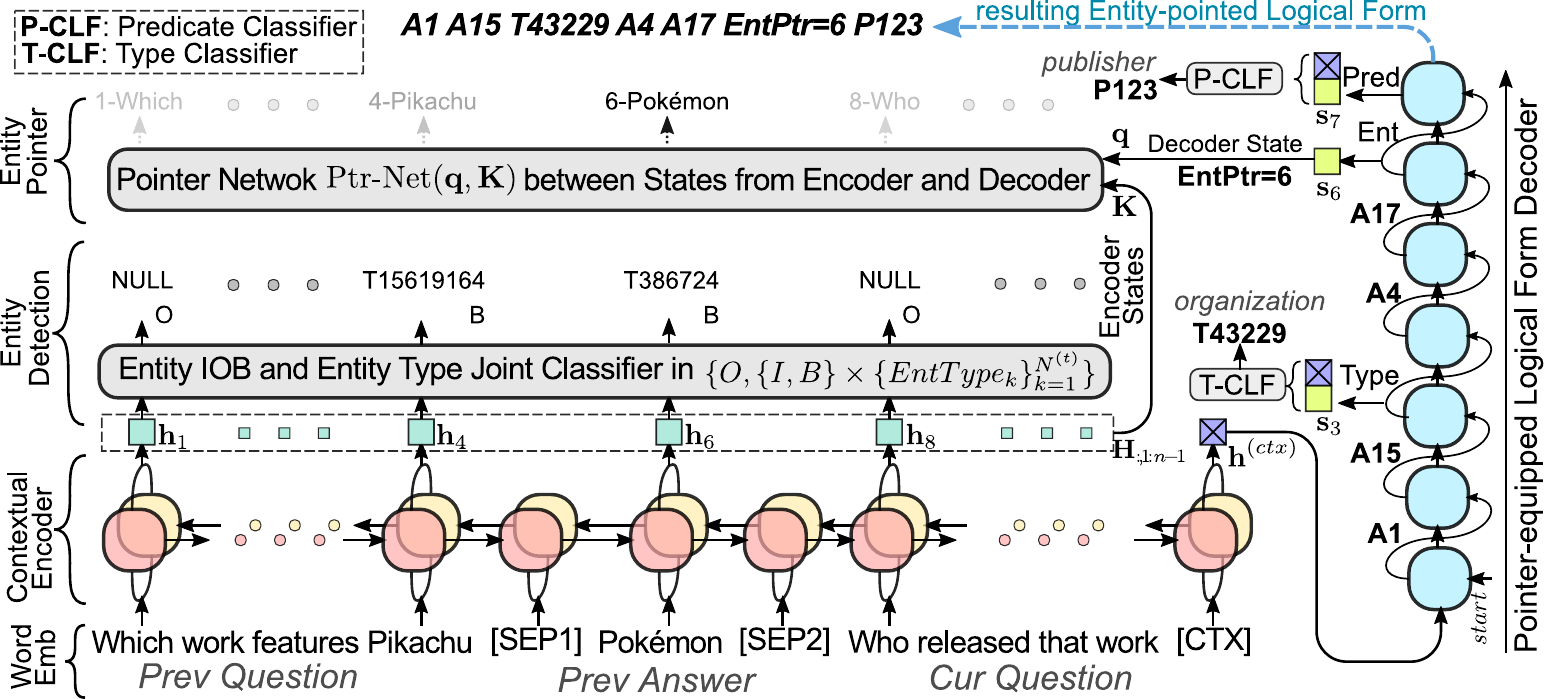}
	\caption{Proposed \textbf{M}ulti-t\textbf{a}sk \textbf{S}emantic \textbf{P}arsing (MaSP) model. Note that P* and T* are predicate and entity type ids in Wikidata where entity type id originally starts with Q but is replaced with T for clear demonstration.}
	\label{fig:app_illu} 
	\centering
\end{figure*}

\section{Approach}

This section begins with a description of grammars and logic forms used in this work. Then, the proposed model is presented, and finally, model's training and inference  are introduced. 

\subsection{Grammar and Logical Form} \label{sec:sp_grammar_lf}

\paragraph{Grammar}
We use similar grammars and logical forms as defined in \citet{d2a}, with minor modification for better adaptation to the CSQA dataset. The grammars are briefly summarized in Table \ref{tb:csqa_grammer_full}, where each operator consists of three components: semantic category, a function name, and a list of arguments with specified  semantic categories. Semantic categories can be classified into two groups here w.r.t. the ways for instantiation: one is referred to as \emph{entry} semantic category (i.e., $\{e, p, tp, u\_num\}$ for entities, predicates, types, numbers) whose instantiations are  constants parsed from a question, and another is referred to as \emph{intermediate} semantic category (i.e., $\{start, set, num, bool\}$) whose instantiation is the output of an operator execution.

\paragraph{Logical Form} 
A KB-executable logical form is intrinsically formatted as an ordered tree where the root is the semantic category $start$, each child node is constrained by the nonterminal (i.e., the un-instantiated semantic category in parenthesis) of its parent operator, and leaf nodes are instantiated entry semantic categories, i.e., constants. 

To make the best of well-performed sequence to sequence (seq2seq) models \cite{vaswani2017attention,bahdanau2015neural} as a base for semantic parsing, we represent a tree-structured logical form as a sequence of operators and constants via depth-first traversal over the tree. Note, given guidance of grammars, we can recover corresponding tree structure from a sequence-formatted logical form.

\subsection{Proposed Model}

The structure of our proposed \textbf{M}ulti-t\textbf{a}sk \textbf{S}mantic \textbf{P}arsing (MaSP) model is illustrated in Figure \ref{fig:app_illu}. The model consists of four components: i.e., word embedding, contextual encoder, entity detection and pointer-equipped logical form decoder. 

\subsubsection{Embedding and Contextual Encoder}

To handle ellipsis or coreference in conversations, our model takes current user question combined with dialog history as the input question $U$. In particular, all those sentences are concatenated with a \emph{[SEP]} separated, and then a special token \emph{[CTX]} is appended. 
We apply wordpiece tokenizing \cite{wu2016google} method, and then use a word embedding method \cite{mikolov2013distributed} to transform the tokenized question to a sequence of  low-dimension distributed embeddings, i.e., $\bm{X} = [\bm{x_1}, \cdots, \bm{x_n}] \in\mathbb R^{d_e \times n}$ where $d_e$ denotes embedding size and $n$ denotes question length. 

Given word embeddings $\bm{X}$, we use stacked two-layer multi-head attention mechanism in the Transformer \cite{vaswani2017attention} with learnable positional encodings as an encoder to model contextual dependencies between tokens, which results in context-aware representations $\bm{H} =  [\bm{h_1}, \cdots, \bm{h_n}] \in\mathbb R^{d_e \times n}$. And, contextual embedding for token \emph{[CTX]} is used as the semantic representation for entire question, i.e., $\bm{h^{(ctx)}} \triangleq \bm{h_n}$

\subsubsection{Pointer-Equipped Decoder}\label{sec:app_seq2seq_dec}

Given contextual embeddings $\bm{H}$ of a question, we employ stacked two-layer masked attention mechanism in  \cite{vaswani2017attention} as the decoder to produce sequence-formatted logical forms. 

In each decoding step, the model first predicts a token from a small decoding vocabulary $\mathbb{V}^{(dec)}$ = \{$start$, $end$, $e$, $p$, $tp$, $u\_num$, A1, $\cdots$, A21\} , where $start$ and $end$ indicate the start and end of decoding, $A1, \cdots, A21$ are defined in Table \ref{tb:csqa_grammer_full}, and $e$, $p$, $tp$ and $u\_num$ denote entity, predicate, type and number entries respectively. A neural classifier is established to predict current decoding token, which is formally denoted as 
\begin{align} \label{eq:tk}
\bm{p_j^{(tk)}} \!\!=\! \softmax(\ffn\nolimits(\bm{s_j}; {\theta^{(tk)}})),
\end{align}
where $\bm{s_j}$ is decoding hidden state of current (i.e., $j$-th) step, $\ffn\nolimits(\cdot; \theta)$ denotes a $\theta$-parameterized two-layer feed forward network with an activation function inside, and $\bm{p_j^{(tk)}} \!\!\in\!\!\mathbb{R}^{|\mathbb{V}^{(dec)}|}$ is a predicted distribution over $\mathbb{V}^{(dec)}$ to score candidates\footnote{Superscript in bracket denotes the type instead of index.}. 

Then, a $\ffn(\cdot)$ or a pointer network \cite{vinyals2015pointer} is utilized to predict instantiation for entry semantic category (i.e., $e$, $p$, $tp$ or $u\_num$ in $\mathbb{V}^{(vec)}$) if it is necessary. 
\begin{itemize}
	\item 
	For predicate $p$ and type $tp$, two parameter-untied $\ffn(\cdot)$ are used as
	\begin{align}
	\bm{p_j^{(p)}}  \!\!\!=\!\!  \softmax(\ffn\nolimits([\bm{s_j};\! \bm{h^{(ctx)}}]; \!{\theta^{(p)}}\!)),
	\end{align}
	\begin{align}
	\bm{p_j^{(t)}} \!\!\!=\!\!  \softmax(\ffn\nolimits([\bm{s_j};\! \bm{h^{(ctx)}}]; \!{\theta^{(t)}}\!)),
	\end{align}
	where $\bm{h^{(ctx)}}$ is semantic embedding of entire question, $\bm{s_j}$ is current hidden state, $\bm{p_j^{(p)}}\!\!\in\!\mathbb{R}^{N^{(p)}}$ and $\bm{p_j^{(t)}}\!\!\in\!\mathbb{R}^{N^{(t)}}$ are predicted distributions over the predicate and type instantiation candidates respectively, and $N^{(p)}$ and $N^{(t)}$ are the numbers of distinct predicates and types in the knowledge base.
	\item 
	For entity $e$ and number $u\_num$, two parameter-untied pointer-networks \cite{vinyals2015pointer} with learnable bilinear layer are employed to point toward the targeted entity\footnote{Toward the first one if entity consists of multiple words.} and number, which are defined as follows.
	\begin{align}
	\bm{p_j^{(e)}}  = \softmax(\bm{s_j} ^T \bm{W^{(e)}} \bm{H}_{:,1:n-1}),
	\end{align}
	\begin{align}
	\bm{p_j^{(n)}}  = \softmax(\bm{s_j} ^T \bm{W^{(n)}} \bm{H}_{:,1:n-1}),
	\end{align}
	where $\bm{H}_{:,1:n-1}$ is contextual embedding of tokens in the question except \emph{[CTX]}, $\bm{W^{(e)}}$ and $\bm{W^{(n)}}$ are weights of pointer-network for entity and number, $\bm{p_j^{(e)}}, \bm{p_j^{(n)}} \in \mathbb{R}^{n-1}$ are the resulting distributions over positions of input question, and $n$ is the length of the question. 
\end{itemize}

The pointer network is also used for semantic parsing in \cite{jia2016recombination}, where the pointer aims at copying out-of-vocabulary words from a question over small-scale KB. Different from that, the pointer used here aims at locating the targeted entity and number in a question, which has two advantages. First, it handles the coreference problem by considering the context of entity mentions in the question. Second, it solves the problem caused by huge entity vocabulary, which reduces the size of decoding vocabulary from several million (i.e., the number of entities in KB) to several dozen (i.e., the length of the question).

\subsubsection{Entity Detection and Linking}\label{sec:app_joint}
\begin{figure}[htbp] 
	\centering
	\includegraphics[width=0.48\textwidth]{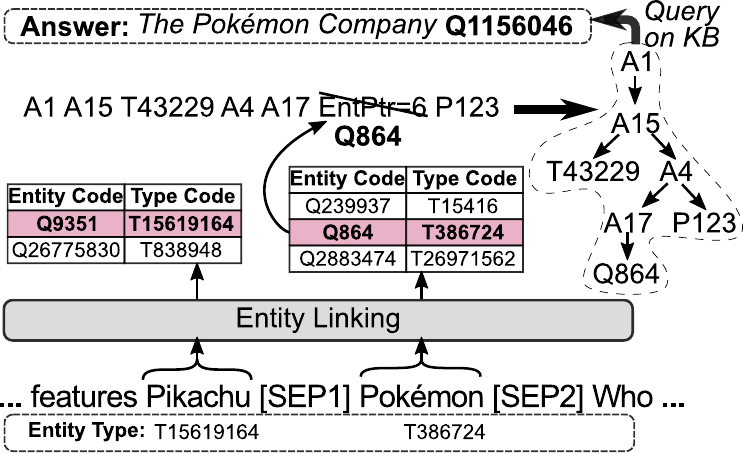}
	\caption{Transformation from entity-pointed logical form to KB-executable logical form for KB querying.}
	\label{fig:app_edl_infer} 
	\centering
\end{figure}

To map the pointed positions to entities in KB, our model also detects entity mentions for the input question, as shown as the ``Entity Detection" part of Figure \ref{fig:app_illu}. 

We observe that multiple entities in a large-scale KB usually have same entity text but different types, leading to named entity ambiguity. Therefore, we design a novel type-aware entity detection module in which the prediction is fulfilled in a joint space of IOB tagging and corresponding entity type for disambiguation. Particularly, the prediction space is defined as $\mathbb{E} = \{ O, \{I, B\}\times \{ET_k\}_{k=1}^{N^{(t)}}\}$ where $ET_k$ stands for the $k$-th entity type label, $N^{(t)}$ denotes number of distinct entity types in KB, and $|\mathbb{E}| = 2 \times N^{(t)} + 1$. 

The prediction for both entity IOB tagging and entity type is formulated as 
\begin{align}\label{eq:ed}
\bm{p_i^{(ed)}}\!\!\!=\!\! \softmax(\ffn\nolimits(\bm{h_i};\!{\theta^{(ed)}}\!)\!),\!\!~\forall i \!\in\! [1, \!n\!\!-\!\!1]
\end{align}
where $\bm{h_i}$ is the contextual embedding of the $i$-th token in the question, and $\bm{p_i^{(ed)}} \in\mathbb{R}^{|\mathbb{E}|}$ is the predicted distribution over $\mathbb{E}$. 

Given the predicted IOB labels and entity types, we take the following steps for entity linking. First, the predicted IOB labels are used to locate all entities in the question and return corresponding entity mentions. Second, an inverted index built on the KB is leveraged to find entity candidates in KB based on each entity mention. Third, the jointly predicted entity types are used to filter out the candidates with unwanted types, and the remaining entity with the highest inverted index score is selected to substitute the pointer. This process is shown as the bottom part of Figure \ref{fig:app_edl_infer}. 

During inference phase, the final logical form is derived by replacing entity pointers in entity-pointed logical form from \S\ref{sec:app_seq2seq_dec} with entity linking results, and is then executed on the KB to retrieve an answer for the question, as shown as the top part of Figure \ref{fig:app_edl_infer}.

\subsection{Learning and Inference} \label{sec:app_learn_infer}

\paragraph{Model Learning}
During the training phase, we first search gold logical forms for questions in training data over KB if only weak supervision is provided. Then we conduct multi-task learning for semantic parsing and entity detection. The final loss is defined as 
\begin{align} \label{eq:loss}
L = \alpha L^{(sp)}  + L^{(ed)},
\end{align}
where $\alpha>0$ is a hyperparameter for a tradeoff between semantic parsing and entity detection, and $L^{(sp)}$ and $L^{(ed)}$ are negative log-likelihood losses of semantic parsing and entity detection defined as follows.
\begin{align}
&L^{(sp)} \! = \! \!  - \dfrac{1}{|\mathcal{D}|}\! \sum_{\mathcal{D}}\!\dfrac{1}{m}\sum_{j=1}^{m} \log \bm{p_j^{(tk)}}\!\!~_{[tk'=y_j^{(tk)}]}  \\
\notag &~~+ \sum_{c \in \{p, t, e, n\}} I_{(y_j^{(tk)}=c)} \log  \bm{p_j^{(c)}}\!\!~_{[c'=y_j^{(c)}]} \\
&L^{(ed)} \!= \! \! - \dfrac{1}{|\mathcal{D}|}\! \sum_{\mathcal{D}}\!\!\dfrac{1}{n\! \! -\! \! 1}\!\!\sum_{i=1}^{n-1}\!\log\!\bm{p_i^{(ed)}} \!\!\!\!~_{[ed'=y_i^{(ed)}]}
\end{align}
In the two equations above, $y_j^{(tk)}$ is gold label for decoding token in $\mathbb{V}^{(dec)}$; $y_j^{(p)}, y_j^{(t)}, y_j^{(e)}~\text{and}~y_j^{(n)}$ are gold labels for predicate, type, entity position and number position for instantiation; $\bm{p_j^{(tk)}}$, $[\bm{p_j^{(c)}}]_{c \in\{p, t, e, n\}}$, and $\bm{p_j^{(ed)}}$ are defined in Eq.(\ref{eq:tk}-\ref{eq:ed}) respectively; and $m$ denotes the decoding length. 

Here, we use a single model to handle two subtasks simultaneously, i.e., semantic parsing and entity detection. This multi-task learning framework enables each subtask to leverage supervision signals from the others, and thus improves the final performance for KB-QA.

\paragraph{Grammar-Guided Inference} 
The grammars defined in Table \ref{tb:csqa_grammer_full} are utilized to filter illegal operators out in each decoding step. An operator is legitimate if its left-hand semantic category in the definition is identical to the leftmost nonterminal (i.e., un-instantiated semantic category) in the incomplete logical form parsed so far.
In particular, the decoding of a logical form begins with the semantic category $start$. During decoding, the proposed semantic parsing model recursively rewrites the leftmost nonterminal in the logical form by 1) applying a legitimate operator for an intermediate semantic category, or 2) instantiation for one of entity, predicate, type or number for an entry semantic category. The decoding process for the parsing terminates until no nonterminals remain. 

Furthermore, beam search is also incorporated to boost the performance of the proposed model during the decoding. 
And, the early stage execution is performed to filter out illegal logical forms that lead to empty intermediate result. 

\section{Experiments}

\begin{table*}[htbp]\small
	\centering
	\begin{tabular}{l|c|c|c|c|l}
		\hline
		\multicolumn{2}{l|}{\textbf{Methods}}  & HRED+KVmem      & D2A (Baseline)     & MaSP (Ours)  & $\Delta$   \\ \hline
		\textbf{Question Type}   & \textbf{\#Example}     & F1 Score      & F1 Score      & F1 Score    & \\ \hline
		Overall &   203k &9.39\% &  66.70\%& 79.26\% &+12.56\%\\
		Clarification& 9k & 16.35\%  &35.53\%  & 80.79\% &+45.26\%\\
		Comparative Reasoning (All) &15k  & 2.96\%  &48.85\%  & 68.90\% &+20.05\%\\
		Logical Reasoning (All)&22k  &8.33\%  &67.31\%  & 69.04\% &+1.73\%\\
		Quantitative Reasoning (All) &9k & 0.96\%  &56.41\%  & 73.75\% &+17.34\%\\
		Simple Question (Coreferenced)&55k  & 7.26\% &57.69\%  & 76.47\% &+18.78\%\\
		Simple Question (Direct) &82k  & 13.64\%  &78.42\%  & 85.18\% &+6.76\%\\
		Simple Question (Ellipsis) &10k & 9.95\%  & 81.14\% & 83.73\% &+2.59\%\\ \hline
		\textbf{Question Type}  &\textbf{\#Example}    & Accuracy &Accuracy& Accuracy&  \\ \hline
		Verification (Boolean)   &27k     & {21.04\%} &  {45.05\%}        & 60.63\% &+15.58\%        \\
		Quantitative Reasoning (Count)  &24k   & {12.13\%} & {40.94\%}  &43.39\% &+2.45\%           \\
		Comparative Reasoning (Count) &15k  & {8.67\%} & {17.78\%}  & 22.26\% &+4.48\%          \\ \hline
	\end{tabular}
	\caption{Comparisons with baselines on CSQA. The last column consists of differences between MaSP and D2A.}
	\label{tb:csqa_res}
\end{table*}

\subsection{Experimental Settings} \label{sec:exp_setting}
\paragraph{Dataset}
We evaluated the proposed approach on Complex Sequential Question Answering (CSQA) dataset\footnote{\url{https://amritasaha1812.github.io/CSQA}} \cite{csqa}, which is the largest dataset for conversational question answering over large-scale KB. It consists of about 1.6M question-answer pairs in  $\sim$200K dialogs, where 152K/16K/28K  dialogs are used for train/dev/test. Questions are classified as different types, e.g., simple, comparative reasoning, logical reasoning questions. Its KB is built on Wikidata\footnote{\url{https://www.wikidata.org}} in a form of (subject, predicate, object), and consists of 21.2M triplets over 12.8M entities, 3,054 distinct entity types, and 567 distinct predicates. 

\paragraph{Training Setups}
We leveraged a BFS method to search valid logical forms for questions in training data. The buffer size in BFS is set to 1000. Both embedding and hidden sizes in the model are set to $300D$, and no pretrained embeddings are loaded for initialization, and the positional encodings are randomly initialized and learnable. The head number of multi-head attention is $6$ and activation function inside $\ffn(\cdot)$ is $\gelu(\cdot)$ \cite{hendrycks2016gaussian}. 
We used Adam \cite{kingma2014adam} to optimize the loss function defined in Eq.(\ref{eq:loss}) where $\alpha$ is set to $1.5$, and learning rate is set to $10^{-4}$. The training batch size is $128$ for $6$ epochs. And we also employed learning rate warmup within the first $1\%$ steps and linear decay within the rest. The source codes are available at \url{https://github.com/taoshen58/MaSP}.
More details of our implementation are described in Appendix \ref{sec:app_model_details}

\paragraph{Evaluation Metrics}

We used the same evaluation metrics as \citet{csqa} and \citet{d2a}.  F1 score (i.e., precision and recall) is used to evaluate the question whose answer is comprised of entities, and accuracy is used to measure the question whose answer type is boolean or number. 

\paragraph{Baselines}
There are few works targeting conversational question answering over a large-scale knowledge base. HRED+KVmem \cite{csqa} and D2A \cite{d2a} are two typical approaches, and we compared them with our proposed approach. Particularly, HRED+KVmem is a memory network \cite{sukhbaatar2015end,lizheng2017end} based seq2seq model, which combines HRED model \cite{serban2016building} with key-value memory network \cite{miller2016key}. D2A\footnote{Overall score of D2A reported in this paper is superior to that in the original paper since our re-implemented grammars for CSQA achieve a better balance between the simple and non-simple question types. For rational and fair comparisons, we report re-run results for D2A in this paper. } is a memory augmented neural symbolic model for semantic parsing in KB-QA, which introduces dialog memory manager to handle ellipsis and co-reference problems in conversations. 

\subsection{Model Comparisons} \label{sec:exp_model_comp}

We compared\footnote{Mores details of comparisons are listed in Appendix \ref{sec:app_more_results}.} our approach (denoted as MaSP) with HRED+KVmem and D2A in Table \ref{tb:csqa_res}. As shown in the table, the semantic parsing based D2A significantly outperforms the memory network based text generation approach (HRED+KVmem), which thus poses a strong baseline. Further, our proposed approach (MaSP) achieves a new state-of-the-art performance, where the overall F1 score is improved by $\sim$12\%. Besides, the improvement is consistent for all question types, which ranges from 2\% to 45\%. 

There are two possible reasons for this significant improvement. First, our approach predicts entities more accurately, where the accuracy of entities in final logical forms increases from 55\% to 72\% compared with D2A. Second, the proposed pointer-equipped logical form decoder in the multi-task learning framework handles coreference better. For instance, given an user question with history, ``\textit{What is the parent organization of that one? // Did you mean Polydor Records ? // No, I meant Deram Records. Could you tell me the answer for that?}'' with coreference, D2A produces ``(\textit{find} \{\textsl{Polydor Records}\}, \textsl{owned by})'' and in contrast our approach produces ``(\textit{find} \{\textsl{Deram Records}\}, \textsl{owned by})''. This also explains the substantial improvement for \textit{Simple Question (Coreferenced)} and \textit{Clarification}\footnote{In CSQA, the performance of \textit{Clarification} closely depends on F1 score for next question, 88\% of which belong to \textit{Simple Question(Coreference)} .}.

We also observed that the improvement of MaSP over D2A for some question types is relatively small, e.g., 1.73\% for logical reasoning questions. A possible reason is that there are usually more than one entities are needed to compose the correct logical form for logical reasoning questions, and our current model is too shallow to parse the multiple entities.
Hence, we adopted deeper model and employed BERT \cite{devlin2018bert} as the encoder (latter in \S\ref{sec:exp_deeper_model}), and found that the performance of logical reasoning questions is improved by 10\% compared to D2A.

\subsection{Ablation Study}
\begin{table}[htbp] \small
	\centering
	\setlength{\tabcolsep}{3.2pt}
	\begin{tabular}{l|c|c|c|c}
		\hline
		\textbf{Methods}     & {Ours}   & {w/o ET}      & {w/o Multi}     & {w/o Both}      \\ \hline
		\textbf{Question Type}    & F1      & F1      & F1      & F1     \\ \hline
		Overall&79.26\% &70.42\% &76.73\% &68.22\% \\
		Clarification&80.79\% &68.01\% &66.30\% &54.64\% \\
		Comparative&68.90\% &66.35\% &61.12\% &58.04\% \\
		Logical&69.04\% &62.63\% &67.81\% &62.51\% \\
		Quantitative&73.75\% &73.75\% &64.56\% &64.55\% \\
		Simple (Co-ref)&76.47\% &64.94\% &74.35\% &63.15\% \\
		Simple (Direct)&85.18\% &75.24\% &84.93\% &75.19\% \\
		Simple (Ellipsis)&83.73\% &78.45\% &82.66\% &77.44\% \\ \hline
		\textbf{Question Type}  & {Accu}  & {Accu} & {Accu} & {Accu} \\ \hline
		Verification & {60.63\%}  & {45.40\%} &  {60.43\%}        & {45.02\%}          \\
		Quantitative& {43.39\%} & {39.70\%} & {37.84\%}  &{43.39\%}           \\
		Comparative& {22.26\%}  & {19.08\%} &  {18.24\%}  & {22.26\%}          \\ \hline
	\end{tabular}
	\caption{ Ablation study. ``w/o ET" stands for removing entity type prediction in Entity Detection of \S\ref{sec:app_joint}; ``w/o Multi" stands for learning two subtasks separately in our framework; and ``w/o Both" stands for a combination of ``w/o ET" and ``w/o Multi".}
	\label{tb:csqa_ablation}
\end{table}
There are two aspects leading to performance improvement, i.e., predicting entity type in entity detection to filter candidates, and multi-task learning framework. We conducted an ablation study in Table \ref{tb:csqa_ablation} for in-depth understanding of their effects. 
\paragraph{Effect of Entity Type Prediction (w/o ET)}
\begin{figure}[t] 
	\centering
	\includegraphics[width=0.4\textwidth]{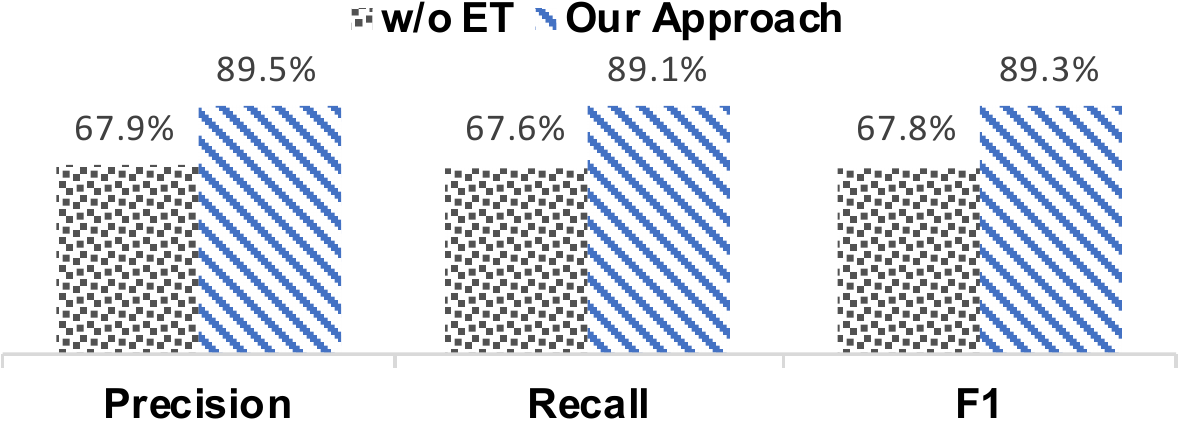}
	\caption{Performance of entity linking. ``w/o ET'' denotes removing entity type filtering. }  
	\label{fig:edl} 
	\centering
\end{figure}

First, the entity type prediction was removed from the entity detection task, which results in 9\% drop of overall F1 score.
We argue that the performance of the KB-QA task is in line with that of entity linking. Hence, we separately evaluated the entity linking task on the test set. As illustrated in Figure \ref{fig:edl}, both precision and recall of entity linking drop significantly without filtering the entity linking results w.r.t. the predicted entity type, which verifies our hypothesis above. 

\paragraph{Effect of Multi-Task Learning (w/o Multi)}

\begin{table}[t] 
	\centering
	\begin{tabular}{l|c|c} \hline
		Accuracy & Ours & w/o Multi \\ \hline
		Entity pointer & 79.8\% & 79.3\% \\
		Predicate & 96.9\% & 96.3\% \\
		Type & 86.8\% & 84.1\% \\
		Number & 89.1\% & 88.3\% \\
		Operators& 79.4\% & 78.7\% \\ 
		\hline 
	\end{tabular} 
	\caption{ Prediction accuracy on each component composing the pointer-equipped logical form.}
	\label{tb:sketch_multi_ablation}
\end{table}

Second, to measure the effect of multi-task learning, we evaluated the KB-QA task when the two subtasks, i.e., pointer-equipped semantic parsing and entity detection, are learned separately.  As shown in Table \ref{tb:csqa_ablation}, the F1 score for every question type consistently drops in the range of 3\% to 14\% compared with that with multi-task learning. We further evaluated the effect of multi-task learning on each subtask. As shown in Table \ref{tb:sketch_multi_ablation}, the accuracy for each component of the pointer-equipped logical form drops with separate learning. 
Meanwhile, we found 0.1\% F1 score reduction (99.4\% vs. 99.5\%) for entity detection subtask compared to the model without multi-task learning, which only poses a negligible effect on the downstream task. 
To sum up, the multi-task learning framework increases the accuracy of the pointer-based logical form generation while keeping a satisfactory performance of entity detection, and consequently improves the final question answering performance.

Note that, considering a combination of removing the entity type filter and learning two subtasks separately (i.e., w/o Both in Table \ref{tb:csqa_ablation}), the proposed framework will degenerate to a model that is similar to Coarse-to-Fine semantic parsing model, another state-of-the-art KB-QA model over small-scale KB \cite{dong2018}. Therefore, an improvement of 11\% of F1 score also verifies the advantage of our proposed framework. 

\subsection{Model Setting Analysis} \label{sec:exp_deeper_model}

As introduced in \S\ref{sec:exp_setting} and evaluated in \S\ref{sec:exp_model_comp}, the proposed framework is built on a relatively shallow neural network, i.e., stacked two-layer multi-head attention, which might limit its representative ability. Hence, in this section, we further exploited the performance of the proposed framework by applying more sophisticated strategies.

\begin{table}[t] \small
	\centering
	\setlength{\tabcolsep}{4pt}
	\begin{tabular}{l|c|c|c}
		\hline
		\textbf{Methods}   & {Vanilla}      & {w/ BERT}     & {w/ Large Beam}      \\ \hline
		\textbf{Question Type}      & F1      & F1      & F1     \\ \hline
		Overall&79.26\% &80.60\% &81.55\% \\
		Clarification&80.79\% &79.46\% &83.37\% \\
		Comparative&68.90\% &65.99\% &69.34\% \\
		Logical&69.04\% &77.53\% &69.41\% \\
		Quantitative&73.75\% &70.43\% &73.75\% \\
		Simple (Co-ref)&76.47\% &77.95\% &79.03\% \\
		Simple (Direct)&85.18\% &86.40\% &88.28\% \\
		Simple (Ellipsis)&83.73\% &84.82\% &86.96\% \\
		\hline
		\textbf{Question Type}      & {Accuracy} & {Accuracy} & {Accuracy} \\ \hline
		Verification& {60.63\%} &  {63.85\%}        & {61.96\%}          \\
		Quantitative& {43.39\%} &  {47.14\%}  & {44.22\%}           \\
		Comparative& {22.26\%} &  {25.28\%}  & {22.70\%}          \\ \hline
	\end{tabular}
	\caption{Comparisons with different experimental settings. ``Vanilla" stands for standard settings of our framework, i.e, MaSP. ``w/ BERT" stands for incorporating BERT. And ``w/ Large Beam" stands for increasing beam search size from 4 to 8.}
	\label{tb:other_settings}
\end{table}

As shown in Table \ref{tb:other_settings}, we first replaced the encoder with pre-trained BERT base model \cite{devlin2018bert} and fine-tuned parameters during the training phase, which results in 1.3\% F1 score improvement over the vanilla one. Second, we increased beam search size from 4 to 8 during the decoding in the inference phase for the standard settings, which leads to 2.3\% F1 score increase. 

\subsection{Error Analysis}
We randomly sampled 100 examples with wrong logical forms or incorrect answers to conduct an error analysis, and found that the errors mainly fall into the following categories. 

\paragraph{Entity Ambiguity} 
Leveraging entity type as a filter in entity linking significantly reduces errors caused by entity ambiguity, but it is still possible that different entities with same text belong to the same type, due to coarse granularity of the entity type, which results in filtering invalidity. For example, it is difficult to distinguish between two persons whose names are both \textit{Bill Woods}. 

\paragraph{Wrong Predicted Logical Form} 
The predicted components (e.g., operators, predicates and types) composing the logical form would be inaccurate, leading to a wrong answer to the question or an un-executable logical form. 

\paragraph{Spurious Logical Form} 
We took a BFS method to search gold logical forms for questions in training set, which inevitably generates spurious (incorrect but leading to correct answers coincidentally) logical forms as training signals. Take the question ``\textit{Which sexes do King Harold, Queen Lillian and Arthur Pendragon possess}" as an example, a spurious logical form only retrieves the genders of ``\textit{King Harold}" and ``\textit{Queen Lillian}", while it gets correct answers for the question. Spurious logical forms accidentally introduce noises into training data and thus negatively affect the performance of KB-QA.

\section{Related Work}
Our work is aligned with semantic parsing based approach for KB-QA. Traditional semantic parsing systems typically learn a lexicon-based parser and a scoring model to construct a logical form given a natural language question \cite{zettlemoyer2007online,wong2007learning,zettlemoyer2009learning,kwiatkowski2011lexical,andreas2013semantic,artzi2013weakly,zhao2014type,long2016simpler}. For example, \citet{zettlemoyer2009learning} and \citet{artzi2013weakly} learn a CCG parser, and \citet{long2016simpler} develop a shift-reduce parser to construct logical forms.

Neural semantic parsing approaches have been gaining rising attention in recent years, eschewing the need for extensive feature engineering \cite{jia2016recombination,ling2016latent,xiao2016sequence}. Some efforts have been made to utilize the syntax of logical forms \cite{rabinovich2017abstract,krishnamurthy2017neural,cheng2017learning,yin2017syntactic}. For example, \citet{dong2016} and \citet{alvarez2016tree} leverage an attention-based encoder-decoder framework to translate a natural language question to tree-structured logical form. 

Recently, to handle huge entity vocabulary existing in a large-scale knowledge base, many works take a stepwise approach. For example, \citet{liang2016neural}, \citet{dong2016}, and \citet{d2a} first process questions using a name entity linking system to find entity candidates, and then learn a model to map a question to a logical form based on the candidates. \citet{dong2018} decompose the task into two stages: first, a sketch of the logical form is predicted, and then a full logical form is generated with considering both the question and the predicted sketch.

Our proposed framework also decomposes the task into multiple subtasks but is different from existing works in several aspects. First, inspired by pointer network \cite{vinyals2015pointer}, we replace entities in a logical form with the starting positions of their mentions in the question, which can be naturally used to handle coreference problem in conversations. Second, the proposed pointer-based semantic parsing model can be intrinsically extended to jointly learn with entity detection for fully leveraging all supervision signals. Third, we alleviate entity ambiguity problem in entity detection \& linking subtask, by incorporating entity type prediction into entity mention IOB labeling to filter out the entities with unwanted types.

\section{Conclusion}
We studied the problem of conversational question answering over a \textit{large-scale} knowledge base, and proposed a multi-task learning framework which learns for type-aware entity detection and pointer-equipped logical form generation simultaneously. The multi-task learning framework takes full advantage of the supervisions from all subtasks, and consequently increases the performance of final KB-QA problem. Experimental results on a large-scale dataset verify the effectiveness of the proposed framework. In the future, we will test our proposed framework on more datasets and investigate potential approaches to handle spurious logical forms for weakly-supervised KB-QA. 

\section*{Acknowledgments}
We acknowledge the support of NVIDIA Corporation and MakeMagic Australia with the donation of GPUs for our research group at University of Technology Sydney. And we also thank anonymous reviewers for their insightful and constructive suggestions. 


\bibliography{emnlp-ijcnlp-2019}
\bibliographystyle{acl_natbib}

\appendix

\section{Model Details}  \label{sec:app_model_details}

\subsection{Word Embedding}  \label{sec:app_word_emb}
Given an user question sentence $U$, a tokenizing method (e.g., punctuation or wordpiece tokenizer \cite{wu2016google}) is applied to the sentence for a list of tokens, i.e., $\bm{U} = [\bm{u_1}, \cdots, \bm{u_{n-1}}, \bm{u'}]$, where $\bm{u_i}$ or $\bm{u'}$ is an one-hot vector whose dimension equals to distinct tokens $N$ in vocabulary, and $n$ is the length of $\bm{U}$. Note that a special token $\bm{u'}$ is appended to the tokenized sentence, corresponding to the token \emph{[CTX]}. Then, randomly initialized or pre-trained \cite{mikolov2013distributed,pennington2014glove} embeddings are applied to $\bm{U}$ and thus transform discrete tokens to a sequence of low-dimension distributed embeddings, i.e.,  $\bm{X} = [\bm{x_1}, \bm{x_2}, ..., \bm{x_n}] \in \mathbb{R}^{d_e\times n}$ where $d_e$ is embedding size. This process is formulated as $\bm{X}  = \bm{W^{(enc)}}\bm{U} $ where $\bm{W^{(enc)}}\in\mathbb{R}^{d_e\times N}$ is the trainable word embedding weight matrix. 

\subsection{Pointer-equipped Semantic Parsing}

\subsubsection{Encoder of Seq2seq Model}
To model contextual dependencies between tokens and generate context-aware representations, we leverage stacked two-layer multi-head attention mechanism with additive positional encoding \cite{vaswani2017attention}. The stacking scheme is identical to that in \cite{vaswani2017attention}: two-layer feed forward network with activation function (FFN) follows each multi-head attention, and residual connection \cite{he2016deep} with layer normalization \cite{lei2016layer} is applied. 
This process is briefly denoted as 
\begin{align}
	\bm{H} &= [\bm{h_1}\!, \cdots, \!\bm{h_n}] \!\triangleq \!\bm{X'} \!\!\in\!\mathbb R^{d_e \times n},\!\text{where}, \\
	^{2\times}[\bm{X'} &= \ffn(\multihead\nolimits(\bm{X'}, \bm{X'}, \bm{X'}))], \\
	\bm{X'} &= \bm{X} + \bm{W^{(pe)}},
\end{align}
where $\bm{H}$ is a sequence of contextual embeddings, $\bm{W^{(pe)}}\in\mathbb{R}^{d_e\times n}$ is learnable weights of PE and the three arguments for $\multihead$ are \emph{value}, \emph{key}, \emph{query} for an attention mechanism. 

\subsubsection{Decoder of Seq2seq Model} 

Similar to token embedding in encoder (\S\ref{sec:app_word_emb}), we embed the $j$-th decoder input token as $\bm{z_j}$ via a randomly initialized embedding weight matrix $\bm{W^{(dec)}}\in\mathbb{R}^{d_e\times |\mathbb{V}^{(dec)}|}$. We use $\bm{Z}=[\bm{z_1}, \cdots, \bm{z_m}] \in\mathbb{R}^{d_e\times m}$ to represent all tokens in a gold logical form sketch, where $m$ denotes the length of gold sketch. 

The basic structure of proposed logical form decoder is same as that in the original Transformer \cite{vaswani2017attention} except only two stacked layers are used here. Each layer of the decoder is bottom-up comprised of self-attention with forward mask, cross attention between decoder and encoder and FFN, which we briefly formulate as
\begin{align}
	\bm{S} &=[\bm{s_1}\!, \cdots, \!\bm{s_m}] \!\triangleq \!\bm{Z} \!\in\!\mathbb R^{d_e \times m},\text{where}, \\
	^{2\times}[ \bm{Z} & =  \ffn(\multihead\nolimits( \\
	\notag  &\bm{H}, \bm{H}, \multihead\nolimits^{mask}(\bm{Z},\bm{Z},\bm{Z})))].
\end{align}
where $\bm{S}$ is a sequence of decoding hidden states. 

\subsection{Multi-task Learning} 

We propose to employ a multi-task learning strategy to learn a entity detection (ED) model jointly with the pointer-equipped semantic parsing model because the supervision information from ED, i.e, IOB tagging, can provide all entities spans in the input question, which thus results in better performance than separate learning. 

The reasons why we use a multi-task learning to jointly learn the semantic parsing model and ED rather than directly equip the semantic parsing model with span prediction \cite{seo2017bidirectional} are that 1) the supervision information of the entities not existing in the gold logical form but appearing in the question is lost; 2) deeper network is required when predicting the end index of the target as shown in \cite{seo2017bidirectional} and 3) the well-solved entity detection method can provide correction for the pointer even with slight deviation during inference phrase, in contrast, span-based model usually leads to error aggregation. 

\subsection{Inverted Index} Based on each entity text in Wikidata, we traversed its substring whose length is not less than that of its full text minus a threshold, and then, we separately calculated Levenshtein Distance between the full text and each substring as a score for the map from the substring to corresponding full text. Since multiple entities could generate identical substring, we kept maps with largest scores and used the maps to build a dictionary for future queries. 

\section{Supplemental Experiment Results} \label{sec:app_more_results}

\subsection{Precision and Recall for Main Paper}
Since we report the F1 score for brief demonstration in the main paper, in this section, we report the corresponding recall and precision detailedly: 
1) as shown in Table \ref{tb:csqa_res_full}, the results of the proposed model compared with baselines are presented;
2) as shown in Table \ref{tb:csqa_ablation_full}, the ablation study is presented; and 
3) as shown in Table \ref{tb:analysis_full}, the performance improvement comparison after sophisticated strategies applied is provided.

\subsection{Comparison to D2A}  

\begin{table}[htbp] \small
	\centering
	\setlength{\tabcolsep}{4pt}
	\begin{tabular}{l|c|c}
		\hline
		\textbf{Question Type}      & \textbf{D2A}      & \textbf{Ours}          \\ \hline
		Simple Question (Direct) & 2.6 & 1.5 \\
		Clarification & 2.7 &1.4 \\
		Simple Question (Coreferenced) &2.7 &1.4 \\
		Quantitative Reasoning (Count) (All) &2.9 &1.5 \\
		Logical Reasoning (All) &2.7 &1.6 \\
		Simple Question (Ellipsis) &2.6 &1.6 \\
		Verification (Boolean) (All) &2.8 &1.4 \\
		Quantitative Reasoning (All) &2.7 &1.4 \\
		Comparative Reasoning (Count) (All) &2.8 &1.4 \\
		Comparative Reasoning (All) &3.0 &1.4 \\ \hline
		\textbf{Overall} &2.9 &1.5 \\
		\hline
	\end{tabular}
	\caption{The averaged number of entity candidates from entity linking.}
	\label{tb:avg_num_of_ent_candidate}
\end{table}

To further demonstrate that the proposed model is superior to the previous D2A model in term of entity linking and logical form generation, we conduct the following comparisons.

First, as shown in Table \ref{tb:avg_num_of_ent_candidate}, the average number of entity candidates in test set from entity linking of the proposed model is $2\times$ less than that of D2A, which means the proposed approach provides the downstream subtask with more accurate entity linking results.

\begin{table}[htbp] \small
	\centering
	\setlength{\tabcolsep}{4pt}
	\begin{tabular}{l|c|c}
		\hline
		\textbf{Question Type}      & \textbf{D2A}      & \textbf{Ours}          \\ \hline
		Simple Question (Direct) &0.8960 &0.9520 \\
		Clarification &0.8281 &0.9323 \\
		Simple Question (Coreferenced) &0.8177 &0.8952 \\
		Quantitative Reasoning (Count) (All) &0.8385 &0.9581 \\
		Logical Reasoning (All) &0.8726 &0.9791 \\
		Simple Question (Ellipsis) &0.9364 &0.9474 \\
		Verification (Boolean) (All) &0.7448 &0.9637 \\
		Quantitative Reasoning (All) &0.9304 &0.9832 \\
		Comparative Reasoning (Count) (All) &0.8165 &0.9863 \\
		Comparative Reasoning (All) &0.8312 &0.9727 \\  \hline
		\textbf{Overall} &0.8499  &0.9475 \\
		\hline
	\end{tabular}
	\caption{Ratio of non-empty logical form.}
	\label{tb:non_empty_lf}
\end{table}

Second, we compare the proposed model with D2A in term of logical form generation where the logical form would be empty due to timeout or illegal logical forms during beam search. As demonstrated in Table \ref{tb:non_empty_lf}, the proposed model obtains less ratio of empty logical form than D2A.

\begin{table}[htbp] \small
	\centering
	\setlength{\tabcolsep}{1pt}
	\begin{tabular}{l|c|c|c}
		\hline
		\textbf{Question Type}      & \textbf{D2A}      & \textbf{Ours}    & \textbf{+BERT}      \\ \hline
		Simple Question (Direct) &0.7967 &0.8519 &0.8664 \\
		Clarification &0.2385 &0.6408 &0.6414 \\
		Simple Question (Coreferenced) &0.5341 &0.7234 &0.7469 \\
		Quantitative Reasoning (Count) (All) &0.5000 &0.6947 &0.7004 \\
		Logical Reasoning (All) &0.3692 &0.0791 &0.3196 \\
		Simple Question (Ellipsis) &0.7533 &0.8843 &0.8878 \\
		Verification (Boolean) (All) &0.1757 &0.5278 &0.5854 \\
		Quantitative Reasoning (All) &0.8913 &0.9792 &0.9911 \\
		Comparative Reasoning (Count) (All) &0.3235 &0.8924 &0.9121 \\
		Comparative Reasoning (All) &0.2483 &0.9053 &0.9242 \\ \hline
		\textbf{Overall} &0.5522 &0.7167 &0.7546 \\
		\hline
	\end{tabular}
	\caption{accuracy of entities in predicted logical form.}
	\label{tb:accu_of_entities_in_logical_form}
\end{table}

Third, we list the accuracies of the entities appearing in the predicted logical form for D2A, our standard approach and BERT-based model, which verifies that the proposed approach can significantly improve the performance of entity linking during entity detection and entity prediction during logical form generation. Note that the analysis for performance reduction of \textit{Logical Reasoning (All)} is elaborated in the main paper.

\subsection{Multi-task Learning}  

The multi-task learning framework increases the accuracy of logical form generation while keeping a satisfactory performance of entity detection, and consequently improves the final question answering task via logical form execution. In this section, we detailedly list all metrics to measure the performance for both two subtasks in the case of our approach with or without  multi-task learning. To evaluate the logical form generation, we also apply BFS method to test set for gold logical form (inevitably existing spurious ones). 

\begin{table}[htbp] \small
	\centering
	\setlength{\tabcolsep}{2pt}
	\begin{tabular}{l|c|c}
		\hline
		\textbf{Question Type}      & \textbf{Ours}      & \textbf{w/o Multi}          \\ \hline
		Comparative Reasoning (All)  & 0.1885 & 0.1885 \\
		Logical Reasoning (All)  & 0.6256 & 0.6188 \\
		Quantitative Reasoning (All)  & 0.6403 & 0.6188 \\
		Simple Question (Coreferenced)  & 0.8721 & 0.8663 \\
		Simple Question (Direct)  & 0.8772 & 0.8715 \\
		Simple Question (Ellipsis)  & 0.9073 & 0.9034 \\
		Comparative Reasoning (Count) (All)  & 0.1601 & 0.1495 \\
		Quantitative Reasoning (Count) (All)  & 0.5711 & 0.5564 \\
		Verification (Boolean) (All)  & 0.7638 & 0.7565 \\ \hline
		\textbf{Overall} & 0.7940 & 0.7872 \\
		\hline
	\end{tabular}
	\caption{Sketch accuracy for logical form generation.}
	\label{tb:multi_task_sketch_accuracy}
\end{table}

\begin{table}[htbp] \small
	\centering
	\begin{tabular}{l|l|c|c}
		\hline
		\multicolumn{2}{l|}{}                    & Ours & w/o Multi \\ \hline
		\multirow{4}{*}{IOB Tagging} & Accuracy  & 0.9967 & 0.9975    \\ \cline{2-4} 
		& F1 Score  & 0.9941 & 0.9955   \\ \cline{2-4} 
		& Precision &  0.9960 & 0.9972  \\ \cline{2-4} 
		& Recall    &  0.9923 & 0.9938   \\ \hline
		\multirow{4}{*}{Entity Type} & Accuracy  &  0.9822 & 0.9844   \\ \cline{2-4} 
		& F1 Score  & 0.9674 & 0.9717  \\ \cline{2-4} 
		& Precision &  0.9958 & 0.9971  \\ \cline{2-4} 
		& Recall    &0.9407 & 0.9475 \\ \hline
	\end{tabular}
	\caption{Performance of IOB tagging and entity type prediction.}
	\label{tb:multi_task_ner}
\end{table}

As shown in Table \ref{tb:multi_task_sketch_accuracy} and \ref{tb:multi_task_ner}, the model with multi-task learning can outperform that without multi-task learning in term of logical form generation from semantic parsing model. And, although $\sim$ 0.002 performance reduction is observed for entity detection subtask, the performance of entity detection and linking is good enough for the downstream task, which thus poses a very minor effect on the performance of KB-QA. 

\subsection{BFS Success Ratio} 

\begin{table}[htbp] \small
	\centering
	\setlength{\tabcolsep}{4pt}
	\begin{tabular}{l|c|c}
		\hline
		\textbf{Question Type}      & \textbf{\#Example}      & Ratio          \\ \hline
		Simple Question (Direct) & 274527 & 0.96 \\
		Simple Question (Ellipsis) & 34549 & 0.97 \\
		Quantitative Reasoning (All) & 58976 & 0.46 \\
		Quantitative Reasoning (Count) (All) & 114074 & 0.67 \\
		Logical Reasoning (All) & 66161 & 0.61 \\
		Simple Question (Coreferenced) & 173765 & 0.86 \\
		Verification (Boolean) (All) & 77167 & 0.75 \\
		Comparative Reasoning (Count) (All) & 59557 & 0.37 \\
		Comparative Reasoning (All) & 57343 & 0.32 \\ \hline
	\end{tabular}
	\caption{The BFS search success ratio w.r.t. difference question type. }
	\label{tb:bfs_success_ratio}
\end{table}

Given the final answer to a question as well as gold entities, predicates and types, we conduct a BFS method to search the gold logical form, which may result in search failure due to limited time and buffer. We list the success ratio of BFS for training data of CSQA in Table \ref{tb:bfs_success_ratio}.

\section{Supplemental Analysis}

We also observe that the improvement of MaSP over D2A for some question types is relatively small especially for logical reasoning questions. Furthermore, for logical reasoning, we find that the accuracy of entities in final logical forms is only 8\%, and there are usually two distinct entities needed to produce a correct logical form. This means the presented shallow network, i.e., two-layer multi-head attention, cannot handle such complex cases. 
We study a case here for better understanding. Given, ``\textit{Which diseases are a sign of lead poisoning or pentachlorophenol exposure?}'', D2A produces ``(\textit{union} (\textit{find} \{\textsl{lead poisoning}\}, \textsl{symptoms}), ({\textsl{pe...ol exposure}}))'' where entities are correct but operator is wrong, our approach produces ``(\textit{union} (\textit{find} \{\textsl{pe...ol exposure}\}, \textsl{symptoms}), (\textit{union} (\textit{find} \{\textsl{pe...ol exposure}\}, \textsl{symptoms}))'' where the entities are wrong, while our approach plus BERT \cite{devlin2018bert} as encoder can produce correct logical form that is ``(\textit{union} (\textit{find} \{\textsl{pe...ol exposure}\}, \textsl{symptoms}), (\textit{union} (\textit{find} \{\textsl{lead poisoning}\}, \textsl{symptoms}))''.


\begin{table*}[htbp]\small
	\centering
	\begin{tabular}{l|c|cc|cc|cc}
		\hline
		\multicolumn{2}{l|}{\textbf{Methods}}  & \multicolumn{2}{c|}{HRED+KVmem}      & \multicolumn{2}{c|}{D2A (Baseline)}     & \multicolumn{2}{c}{Our Approach}      \\ \hline
		\textbf{Question Type}   & \textbf{\#Example}     & Recall       & Precision      & Recall       & Precision      & Recall       & Precision     \\ \hline
		Overall &   - & 18.40\% & 6.30\% & 66.83\% & 66.57\% & 78.07\% & 80.48\% \\
		Clarification&12k & 25.09\% & 12.13\% & 37.24\% & 33.97\% & 84.18\% & 77.66\% \\
		Comparative Reasoning (All) &15k  & 2.11\% & 4.97\% & 44.14\% & 54.68\% & 59.83\% & 81.20\% \\
		Logical Reasoning (All)&22k  & 15.11\% & 5.75\% & 65.82\% & 68.86\% & 61.92\% & 78.00\% \\
		Quantitative Reasoning (All) &9k & 0.91\% & 1.01\% & 52.74\% & 60.63\% & 69.14\% & 79.02\% \\
		Simple Question (Coreferenced)&55k  & 12.67\% & 5.09\% & 58.47\% & 56.94\% & 76.94\% & 76.01\% \\
		Simple Question (Direct) &82k  & 33.30\% & 8.58\% & 79.50\% & 77.37\% & 86.09\% & 84.29\% \\
		Simple Question (Ellipsis) &10k & 17.30\% & 6.98\% & 84.67\% & 77.90\% & 85.50\% & 82.03\% \\ \hline
		\textbf{Question Type}  &\textbf{\#Example}    & \multicolumn{2}{c|}{Accuracy} & \multicolumn{2}{c|}{Accuracy} & \multicolumn{2}{c}{Accuracy} \\ \hline
		Verification (Boolean)   &27k     & \multicolumn{2}{c|}{21.04\%} &  \multicolumn{2}{c|}{45.05\%}        & \multicolumn{2}{c}{60.63\%}          \\
		Quantitative Reasoning (Count)  &24k   & \multicolumn{2}{c|}{12.13\%} &  \multicolumn{2}{c|}{40.94\%}  & \multicolumn{2}{c}{43.39\%}           \\
		Comparative Reasoning (Count) &15k  & \multicolumn{2}{c|}{8.67\%} &  \multicolumn{2}{c|}{17.78\%}  & \multicolumn{2}{c}{22.26\%}          \\ \hline
	\end{tabular}
	\caption{Results of comparisons for KB-QA with baselines.}
	\label{tb:csqa_res_full}
\end{table*}

\begin{table*}[htbp] \small
	\centering
	\setlength{\tabcolsep}{5pt}
	\begin{tabular}{l|cc|cc|cc|cc}
		\hline
		\textbf{Methods}     & \multicolumn{2}{c|}{Our Approach}   & \multicolumn{2}{c|}{w/o ET}      & \multicolumn{2}{c|}{w/o Multi}     & \multicolumn{2}{c}{w/o Both}      \\ \hline
		\textbf{Question Type}    & Recall       & Precision      & Recall       & Precision      & Recall       & Precision      & Recall       & Precision     \\ \hline
		Overall & 78.07\% & 80.48\% & 68.78\% & 72.15\% & 75.75\% & 77.73\% & 66.75\% & 69.75\% \\
		Clarification& 84.18\% & 77.66\% & 69.79\% & 66.32\% & 70.12\% & 62.88\% & 56.96\% & 52.51\% \\
		Comparative Reasoning (All)& 59.83\% & 81.20\%  & 57.48\% & 78.45\% & 53.62\% & 71.06\% & 50.86\% & 67.59\% \\
		Logical Reasoning (All) & 61.92\% & 78.00\% & 54.43\% & 73.73\% & 61.04\% & 76.27\% & 54.16\% & 73.91\% \\
		Quantitative Reasoning (All) & 69.14\% & 79.02\%  & 69.14\% & 79.02\% & 60.86\% & 68.73\% & 60.86\% & 68.72\% \\
		Simple Question (Coreferenced)  & 76.94\% & 76.01  & 64.92\% & 64.96\% & 74.65\% & 74.06\% & 63.06\% & 63.24\% \\
		Simple Question (Direct) & 86.09\% & 84.29\%  & 75.87\% & 74.62\% & 85.88\% & 84.01\% & 75.84\% &74.56\% \\
		Simple Question (Ellipsis) & 85.50\% & 82.03\% & 80.12\% & 76.85\% & 84.28\% & 81.11\% & 78.96\% & 75.97\% \\ \hline
		\textbf{Question Type}  & \multicolumn{2}{c|}{Accuracy}  & \multicolumn{2}{c|}{Accuracy} & \multicolumn{2}{c|}{Accuracy} & \multicolumn{2}{c}{Accuracy} \\ \hline
		Verification (Boolean) & \multicolumn{2}{c|}{60.63\%}  & \multicolumn{2}{c|}{45.40\%} &  \multicolumn{2}{c|}{60.43\%}        & \multicolumn{2}{c}{45.02\%}          \\
		Quantitative Reasoning (Count) & \multicolumn{2}{c|}{43.39\%} & \multicolumn{2}{c|}{39.70\%} & \multicolumn{2}{c|}{37.84\%}  & \multicolumn{2}{c}{43.39\%}           \\
		Comparative Reasoning (Count)  & \multicolumn{2}{c|}{22.26\%}  & \multicolumn{2}{c|}{19.08\%} &  \multicolumn{2}{c|}{18.24\%}  & \multicolumn{2}{c}{22.26\%}          \\ \hline
	\end{tabular}
	\caption{Ablation study. ``w/o ET" stands for removing entity type prediction in Entity Detection; ``w/o Multi" stands for learning two subtasks separately in our framework; and ``w/o Both" stands for a combination of ``w/o ET" and ``w/o Multi".}
	\label{tb:csqa_ablation_full}
\end{table*}

\begin{table*}[htbp] \small
	\centering
	\begin{tabular}{l|cc|cc|cc}
		\hline
		\textbf{Methods}   & \multicolumn{2}{c|}{Vanilla}      & \multicolumn{2}{c|}{w/ BERT}     & \multicolumn{2}{c}{Larger Beam Size}      \\ \hline
		\textbf{Question Type}      & Recall       & Precision      & Recall       & Precision      & Recall       & Precision     \\ \hline
		Overall & 78.07\% & 80.48\% & 79.67\% & 81.56\% & 80.39\% & 82.75\% \\ 
		Clarification & 84.18\% & 77.66\% & 83.24\% & 76.01\% & 86.90\% & 80.11\% \\ 
		Comparative Reasoning (All)  & 59.83\% & 81.20\% & 58.79\% & 75.21\% & 60.25\% & 81.67\% \\ 
		Logical Reasoning (All)  & 61.92\% & 78.00\% & 72.56\% & 83.24\% & 62.16\% & 78.58\% \\ 
		Quantitative Reasoning (All)  & 69.14\% & 79.02\% & 66.91\% & 74.35\% & 69.14\% & 79.02\% \\ 
		Simple Question (Coreferenced)  & 76.94\% & 76.01\% & 78.05\% & 77.85\% & 79.54\% & 78.52\% \\ 
		Simple Question (Direct)  & 86.09\% & 84.29\% & 86.84\% & 85.96\% & 89.26\% & 87.33\% \\ 
		Simple Question (Ellipsis)  & 85.50\% & 82.03\% & 86.38\% & 83.32\% & 88.78\% & 85.22\% \\ \hline
		\textbf{Question Type}      & \multicolumn{2}{c|}{Accuracy} & \multicolumn{2}{c|}{Accuracy} & \multicolumn{2}{c}{Accuracy} \\ \hline
		Verification (Boolean)       & \multicolumn{2}{c|}{60.63\%} &  \multicolumn{2}{c|}{63.85\%}        & \multicolumn{2}{c}{61.96\%}          \\
		Quantitative Reasoning (Count)     & \multicolumn{2}{c|}{43.39\%} &  \multicolumn{2}{c|}{47.14\%}  & \multicolumn{2}{c}{44.22\%}           \\
		Comparative Reasoning (Count)   & \multicolumn{2}{c|}{22.26\%} &  \multicolumn{2}{c|}{25.28\%}  & \multicolumn{2}{c}{22.70\%}          \\ \hline
	\end{tabular}
	\caption{Comparisons with different experimental settings. ``Vanilla" stands for standard settings of our framework. ``w/ BERT" stands for incorporating BERT. ``w/ Large Beam" stands for increasing beam search size from 4 to 8.}
	\label{tb:analysis_full}
\end{table*}

\end{document}